\documentclass[11pt]{article}

\usepackage[final]{acl}

\usepackage{times}
\usepackage{latexsym}

\usepackage[T1]{fontenc}

\usepackage[utf8]{inputenc}

\usepackage{microtype}

\usepackage{inconsolata}

\usepackage{graphicx}

\usepackage{url}
\usepackage{graphicx}
\usepackage[table]{xcolor}
\usepackage{colortbl}
\usepackage{booktabs} 

\usepackage{tikz}
\usepackage{subcaption}

\usepackage{float}
\usepackage{subcaption}
\usepackage{enumitem}
\usepackage{tcolorbox}
\usepackage{lipsum} 

\usepackage{listings}

\usepackage{algorithm}
\usepackage{algpseudocode}
\usepackage{float}

\usepackage{amssymb}
\usepackage{xspace,mfirstuc,tabulary}

\usepackage{makecell}
\usepackage{amsmath}

\lstdefinestyle{mycode}{
  backgroundcolor=\color{gray!10},
  frame=single,
  basicstyle=\ttfamily\small,
  breaklines=true,
  numbers=left,
  numberstyle=\tiny\color{gray},
  keywordstyle=\color{blue},
  commentstyle=\color{green!60!black},
  captionpos=b
}
\lstdefinelanguage{json}{
  basicstyle=\ttfamily\small,
  numbers=left,
  numberstyle=\tiny,
  stepnumber=1,
  numbersep=5pt,
  stringstyle=\color{red},
  showstringspaces=false,
  breaklines=true,
  frame=single,
  moredelim=[l][\color{blue}]{:},
  morestring=[b]"
}

\title{Process Reward Models Meet Planning:\\ Generating Precise and Scalable Datasets for Step-Level
Rewards}

\author{
    Raffaele Pisano \\
    Babelscape \\
    \texttt{pisano@babelscape.com} \\
\And
    Roberto Navigli \\
    Babelscape \& Sapienza University of Rome \\
    \texttt{navigli@diag.uniroma1.it} \\
}

\begin{document}
\maketitle
\begin{abstract}
Process Reward Models (PRMs) have emerged as a powerful tool for providing step-level feedback when evaluating the reasoning of Large Language Models (LLMs), which frequently produce chains of thought (CoTs) containing errors even when the final answer is correct. However, existing PRM datasets remain expensive to construct, prone to annotation errors, and predominantly limited to the mathematical domain.
This work introduces a novel and scalable approach to PRM dataset generation based on planning logical problems expressed in the Planning Domain Definition Language (PDDL). 
Using this method, we generate a corpus of approximately one million reasoning steps across various PDDL domains and use it to train PRMs. Experimental results show that augmenting widely-used PRM training datasets with PDDL-derived data yields substantial improvements in both mathematical and non-mathematical reasoning, as demonstrated across multiple benchmarks.
These findings indicate that planning problems constitute a scalable and effective resource for generating robust, precise, and fine-grained training data for PRMs, going beyond the classical mathematical sources that dominate this field.
\end{abstract}

\section{Introduction}

Recently, LLMs have achieved remarkable progress across a wide range of domains, including mathematics, logic, and program synthesis~\cite{openai_o1, deepseek_r1, Qwen3, Magistral, google2025gemini3}. Models that use CoT reasoning have shown clear advantages over those that do not; however, they still frequently exhibit inconsistencies in their intermediate reasoning steps~\cite{Turpin, PRM800k, ProcessBench, Kambhampati7}. It is not uncommon for a model to produce a correct final answer while the accompanying reasoning contains flawed, illogical, or self-contradictory steps.
These observations highlight the importance of evaluating reasoning at the process level, rather than relying solely on final-answer correctness, in order to better detect and mitigate logical imperfections and improve the robustness of LLM reasoning.

Process Reward Models (PRMs) have gained attention as a promising approach to addressing this challenge~\cite{First-PRM}. 
By assigning step-level rewards, PRMs enable a detailed evaluation of individual steps within CoTs.
However, existing PRMs are trained on datasets that present some limitations.
One of the most notable datasets is PRM800k~\cite{PRM800k}, created through manual annotation of reasoning steps, a process that is time-consuming and difficult to scale.
The dataset also provides limited granularity, distinguishing steps only as good, neutral, or bad, and focuses exclusively on mathematical reasoning.
Annotators were required to satisfy a minimum agreement threshold, leaving room for imperfect or unreliable labels.

To avoid human annotation effort, \citet{Math-Shepherd} introduce the Math-Shepherd dataset, constructed via an LLM-based reward annotation approach. However, this method is computationally expensive and yields only a coarse estimate of step-level correctness. In particular, annotations are restricted to binary labels (good vs. bad) and are confined to mathematical problem domains, without extending to broader forms of logical reasoning. 
These limitations reduce data quality and limit potential improvements for PRMs.

To fill these gaps, we propose a novel framework for PRM dataset generation based on the Planning Domain Definition Language (PDDL), a formal language for representing a subset of reasoning problems, namely planning problems.
Our approach enables automatic, rule-based reward assignment and the creation of large-scale, high-precision reasoning datasets that extend beyond mathematical tasks.

\begin{figure*}
    \centering
    \includegraphics[width=1\linewidth]{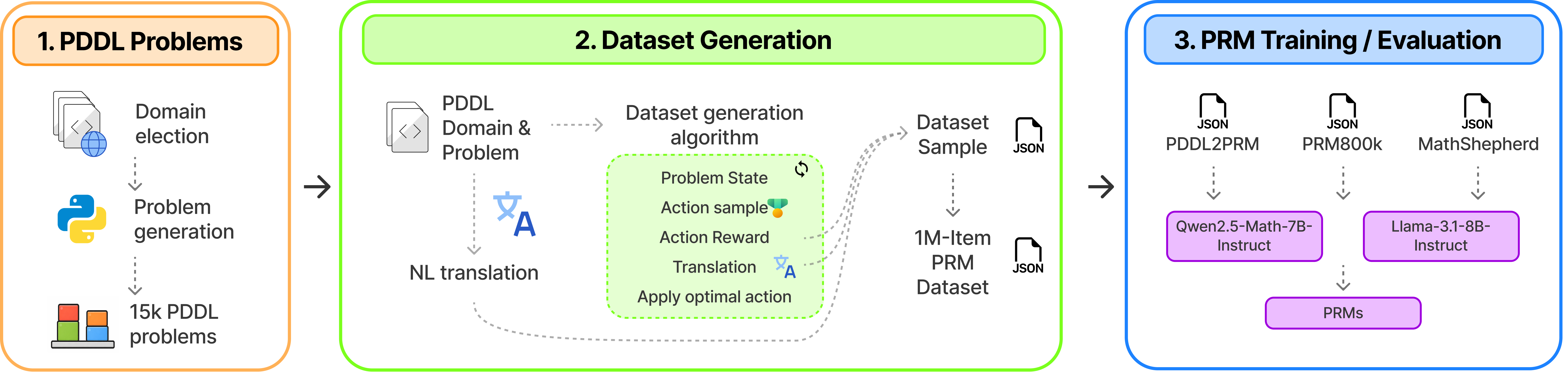}
    \caption{End-to-end workflow of the proposed framework: from PDDL problem generation, to dataset construction with step-level rewards, to PRM training.}
    \label{fig:wf_1}
\end{figure*}

Empirical results show that integrating PDDL-derived data with PRM800k and Math-Shepherd substantially improves PRMs' performance, especially on non-mathematical benchmarks. Our key contributions are:
\begin{itemize}
    \item Introduction of a novel pipeline for PRM dataset generation based on PDDL, providing a scalable and reproducible framework for reasoning supervision.
    \item Generation of a large-scale dataset of around one million reasoning steps across eleven PDDL domains, covering approximately 15,000 planning problems.
    \item Training, evaluation, and release of multiple PRMs with competitive performance, enabling further experimentation and evaluation.
    
    \item Comparison between regression- and classification-based PRM training, showing the advantages of the regression formulation.
    \item Demonstration that PDDL-augmented training improves both accuracy and robustness in step-level reasoning evaluation, extending generalization beyond mathematical domains.
\end{itemize}

Overall, this work establishes PDDL-based dataset generation as a promising direction for developing more structured, interpretable, and scalable reasoning supervision.
The datasets and the trained PRMs used in this work are available at the following link: \url{https://github.com/Babelscape/prm-meets-planning/}.
The whole workflow is shown in Figure \ref{fig:wf_1}.

\section{Related Work}

\subsection{Feedback in Reasoning LLMs}

The notable progress of LLMs in reasoning tasks~\cite{openai_o1, deepseek_r1, Qwen3, Magistral, google2025gemini3}, together with the emergence of Long Chain of Thought~\cite{Towards_Reasoning_Era}, has attracted significant attention to their cognitive and problem-solving capabilities.
Within this context, feedback plays a crucial role, serving as an external signal that can be used to assess and refine CoT-based reasoning.

An important distinction has been drawn between process feedback, which performs step-level evaluation through PRMs, and outcome feedback, which assesses only the final result through Outcome Reward Models  \cite[ORMs]{First-PRM}. PRMs have been shown to be more effective than ORMs in enhancing reasoning performance, particularly in mathematical tasks \cite{PRM800k, PRMvsORM1, PRMvsORM2}. However, outcome-level evaluation is comparatively straightforward to implement, whereas process-level evaluation is inherently more complex.

\subsection{Process Reward Models}
PRMs have been introduced to enable process-level evaluation, focusing on assessing the intermediate reasoning steps rather than only the final answers. However, their effectiveness is constrained by the quality and diversity of the datasets currently available for their training, the creation of which remains a challenging open problem. 
Several approaches have been proposed for generating datasets to support PRM training. \citet{PRM800k} introduced a large-scale dataset in the mathematical domain, where reasoning steps are manually annotated as good, neutral, or bad. While widely adopted, this approach suffers from limited scalability due to the high cost of manual annotation. Moreover, the annotators selected for the final dataset were chosen based on achieving over 75\% agreement with gold labels, implying that up to 25\% of the final annotations may still contain errors. Consequently, while the dataset represents a valuable resource, its reliability is not guaranteed.

The Math-Shepherd dataset \cite{Math-Shepherd}, on the other hand, avoids the need for human annotators by adopting an automatic reward assignment method. For each intermediate step in a reasoning chain, the solution is truncated at that point and multiple continuations are generated using an LLM. The quality of a step is then estimated as the fraction of these continuations that reach the correct final answer. Although this strategy is effective, it remains both computationally expensive and indirect.
As a result, the dataset lacks gold-standard supervision, as it evaluates utility rather than intrinsic correctness, and is ultimately restricted to a binary classification (correct vs. incorrect).

Other approaches, such as PRIME \cite{prime} and FPRWL \cite{fprwl}, bypass explicit annotations by deriving implicit signals from the correctness of final solutions, for example using log-likelihood scores. Efficiency improvements have also been investigated. EPIC-PRM \cite{epicprm} reduces computational costs while maintaining annotation quality. Similarly, \citet{Qwen2.5-PRM} combine the Math-Shepherd strategy with an additional LLM acting as a ``judge'' to double check step quality. 
More recently, \citet{molfese-etal-2026-retraceqa} show that state-of-the-art PRMs struggle to generalize beyond mathematics, particularly to commonsense reasoning tasks. 
Finally, VersaPRM \cite{VersaPRM} extends the scope of PRM dataset generation to domains such as physics, law, and economics. However, it still draws on an annotation procedure that employs an LLM as a step-level judge, making it potentially error-prone, and uses only three labels (good, neutral, bad).

Our work similarly extends the scope of PRM dataset generation beyond the mathematical domain, leveraging logical planning problems across diverse types of task. However, unlike prior approaches, our method assigns rewards using precise, rule-based criteria and provides a richer set of reward levels, thereby addressing key limitations of existing techniques in terms of scalability, precision, and label granularity.

\subsection{PRMs and PDDL}
PDDL is a formal language designed to describe planning domains and problems in a structured and machine-interpretable format. It provides a standardized framework for representing states, actions, and goals, making it a fundamental tool in automated planning and reasoning research.
An increasing number of works have explored the intersection between PDDL and LLMs. Among these, several contributions highlight the potential of planning problems as a testbed for evaluating the reasoning abilities of LLMs. In particular, the works by \citet{kambhampati1, kambhampati2, eval_pddl_capabilities, llm_reasoning_replace_planning} examine the capabilities of LLMs in planning and action reasoning, including their ability to generate and validate simple plans.
\citet{stechly2024cotplanning, kambhampati5} show that LLMs struggle substantially with self-correction when engaged in planning-related tasks, and that the use of CoT offers only limited improvements in their ability to reason through planning problems. In a different direction, \citet{teaching_llms_to_plan} propose a framework in which LLMs are trained not only to output a plan but also to rigorously reason about action applicability, state transitions, and plan validity.
Other approaches \cite{kambhampati3, llms_planning_domain_generators} do not use PDDL to evaluate LLMs; instead, they utilize them to generate PDDL problems that can then be solved by external symbolic planners.
In contrast to these studies, our work neither relies on LLMs to generate PDDL problems nor uses PDDL directly to train or evaluate them. 

\section{PDDL for LLM Reasoning Evaluation}

Our core idea is to exploit the nature of PDDL not only as a source of logical reasoning problems for evaluating reasoning capabilities of LLMs, as previously done in the literature, but also as a structured source for constructing artificial reasoning chains with step-level rewards, which are subsequently used to train PRMs.
The final goal is to create a scalable and low-cost dataset in which each sample includes a PDDL problem expressed in natural-language form, together with a partially solved reasoning trajectory. The trajectory consists of a variable number of intermediate steps, each aligned with a specific action and associated with a precise reward.
The large number of domains and problem instances that can be generated in PDDL naturally enables scalable dataset construction.

\paragraph{Formalization of Planning Concepts}

To clarify the theoretical background underlying our work, we first introduce the fundamental concepts of Classical Planning~\cite{ghallab2004automated}.

A \textit{planning domain} is defined as a tuple:
{
\setlength{\abovedisplayskip}{6pt}
\setlength{\belowdisplayskip}{6pt}
\[
D = \langle S, A, \delta \rangle
\]
}

\noindent where $S$ is a finite set of states, $A$ is a finite set of actions, and 
$\delta : S \times A \rightarrow S$ is a deterministic transition function that specifies the successor state resulting from applying an action in a given state.

A \textit{planning problem} is described by the triple:
\[
P = (D, s_0, G),
\]
where $s_0 \in S$ is the initial state and $G \subseteq S$ denotes the set of goal states to be achieved. 

A \textit{plan} is a finite sequence of actions:
{
\setlength{\abovedisplayskip}{6pt}
\setlength{\belowdisplayskip}{6pt}
\[
\pi = \langle a_1, a_2, \dots, a_n \rangle
\]
}

\noindent such that each transition satisfies $s_{i+1} = \delta(s_i, a_{i+1})$ for $i = 0, \dots, n-1$, and the final state $s_n \in G$.

A \textit{planner} is an algorithm that, given a problem $P$, computes a plan $\pi$ leading from $s_0$ to a state that satisfies the goal condition.

\paragraph{Dataset Generation Framework}

At the core of our approach lies the interpretation of a planning problem as a logical reasoning task, where solving the problem corresponds to finding a sequence of actions that leads from an initial state to a goal state.
From this perspective, the plan represents the reasoning process itself, and each action corresponds to an individual reasoning step.
To generate the dataset we therefore simulate multiple planning steps and assign a reward to each action according to specific evaluation rules.
The following utility functions are introduced to facilitate dataset construction:

\begin{itemize}
    \item $\texttt{get\_rand\_action(P,s)} \rightarrow a$, which returns a random action $a \in A$, drawn from a distribution that includes both applicable and non-applicable actions in state $s \in S$.
    
    \item $\texttt{get\_opt\_action(P,s)} \rightarrow a$, which returns the first action $a \in A$ of the optimal plan starting at state $s$. The optimal plan is computed by an external planner that guarantees optimality (see implementation details in Section~\ref{sec:exp-setup-optimal}).

    \item $\texttt{eval\_action(P,s,a)} \rightarrow r$, which returns a reward $r \in [0,1]$ evaluating the quality of action $a$ in state $s$, based on its executability, proximity to the goal, and whether it belongs to the optimal plan. Further implementation details are provided in Section~\ref{sec:exp-setup-actionrewads}.

    \item $\texttt{translate(P,s,a,r)}$, which converts the action into a natural language reasoning step and adds it to the dataset with reward $r$.
\end{itemize}

Algorithm~\ref{algo:dataset_gen} illustrates the dataset generation process. At each iteration, $y$ random actions are sampled, where $y$ is a domain-dependent parameter controlling how many exploratory steps are generated per state. 
The sampled actions are then evaluated using \texttt{eval\_action(P,s,a)} and translated using \texttt{translate(P,s,a,r)} for inclusion in the dataset. The procedure is repeated until the goal state is reached (i.e. $s \in G$). By sampling diverse candidate actions, including suboptimal, poor, and optimal ones, the algorithm ensures that each state is associated with a rich set of alternative reasoning steps, creating both correct and incorrect trajectories.



\begin{algorithm}[t!]
\caption{\label{algo:dataset_gen}Dataset Generation}
\begin{algorithmic}[1]
\Require $P = ((S, A, \delta), s_0, G)$
\State $s \gets s_0$
\While{$s \notin G$}
    \For{$i = 1$ to $y$}
        \State $a \gets \texttt{get\_rand\_action}(P, s)$
        \State $r \gets \texttt{eval\_action}(P, s, a)$
        \State $\texttt{translate}(P, s, a, r)$
    \EndFor
    \State $a^* \gets \texttt{get\_opt\_action}(P, s)$
    \State $s \gets \delta(s, a^*)$
\EndWhile
\end{algorithmic}
\end{algorithm}

\begin{table*}[ht]
\centering
\resizebox{1.00\textwidth}{!}{
\begin{tabular}{c|l|c}

\toprule
\textbf{Category} & \textbf{Description} & \textbf{Reward}\\
\midrule

Non-executable& $a$ cannot be applied because its preconditions are not satisfied in $S$ & 0.0 \\
Dead-end & $a$ leads to a state $S'$ from which the goal is unreachable & 0.25 \\
Backtracking & $a$ leads to a state $S'$ from which the optimal plan requires revisiting a previous state & 0.5 \\
Suboptimal & $a$ is valid but does not belong to any optimal plan & 0.75 \\
Optimal & $a$ belongs to at least one optimal plan starting from $S$ & 1.0 \\

\bottomrule

\end{tabular}
}
\caption{\label{tab:action_rewards}
Category definitions and reward assignment. Each category is associated with a reward value reflecting the correctness of an action $a$ in a given starting state $S$ and its contribution to the reasoning process.
}
\end{table*}


\section{Experimental Setup}
\subsection{Planning Domain Overview}
We use 11 different PDDL domains, most of which originate from the International Planning Competition ~\cite[IPC]{ipc}, while others are custom-designed by drawing inspiration from existing domains. The domains span a diverse set of planning tasks, which we group into the following categories.

\paragraph{Manipulation and Rearrangement Tasks.}
This category includes domains where an agent rearranges stacked objects to reach a target configuration (\textit{BlocksWorld-3-ops}, \textit{BlocksWorld-4-ops}).

\paragraph{Transportation Tasks.}
These domains involve moving entities across locations optimizing transportation efficiency (\textit{Ferry}, \textit{Logistics}, \textit{Elevator}).

\paragraph{Puzzles and Constraint Satisfaction Tasks.}
This group consists of classical puzzle problems that require planning under strict movement constraints (\textit{Tower of Hanoi}, \textit{N-Puzzle}).

\paragraph{Navigation and Exploration Tasks.}
Domains in this category focus on spatial reasoning and sequential decision-making, often involving irreversible actions (\textit{VisitGrid}, \textit{Sokoban}, \textit{Rooms}, \textit{Spanner}).

Further details are provided in Appendix~\ref{appendix:app_A}.

\subsection{Dataset Generation details}
\subsubsection{PDDL Processing}
\label{sec:exp-setup-optimal}
For handling and manipulating PDDL problems, we rely on the UnifiedPlanning library~\cite{unified_planning_softwarex2025}
which provides a framework to parse PDDL domain and problem files, query the set of applicable actions in each state, and interface with external planners.
The \texttt{get\_opt\_action(P,s)} function, previously introduced, computes the optimal action through the use of the Fast Downward planner~\cite{FastDownward} configured with A* search and the LM-Cut heuristic~\cite{helmert2009landmarks}.
LM-Cut is admissible, meaning that it never overestimates the true cost-to-go to the goal, and therefore guarantees optimal solutions under A* search~\cite{dechter1985optimality}, ensuring that the first action extracted from the plan is optimal.

\subsubsection{Action Reward Assignment}
\label{sec:exp-setup-actionrewads}
To the best of our knowledge, there is no single, universally accepted taxonomy of action types in planning. 
However, several distinctions are useful for the purposes of our work. 
Beyond the basic separation between executable, non-executable, optimal and sub-optimal actions, we identify two additional categories that are particularly relevant.
First, we consider actions that lead to states from which the goal is no longer reachable \cite[dead-end states]{atkins1996detecting}. 
Second, we consider actions that induce backtracking, namely actions that lead to states from which an optimal plan requires revisiting a previously visited state.
Based on these considerations, we adopt a taxonomy of five action categories, each associated with a scalar reward in the range $[0,1]$ reflecting the degree of progress an action provides toward the goal state, as summarized in Table~\ref{tab:action_rewards}.

Accordingly, the previously introduced \texttt{eval\_action(P,s,a)} function is implemented to determine which of the five categories the action \texttt{a} belongs to when executed in state \texttt{s}.
Using the planner and the PDDL processing tools, the function first checks whether the action is \emph{Non-executable} in the current state. If the action is executable, the successor state is computed and assessed to determine whether the planner can still find an optimal plan from it; if this is not possible, the action is labeled as a \emph{Dead-end}. If an optimal plan does exist from the successor state, the function checks whether this plan revisits a previously encountered state, which characterizes \emph{Backtracking}. Finally, if action \texttt{a} belongs to at least one optimal plan, it is labeled as \emph{Optimal}; otherwise, it is classified as \emph{Suboptimal}.

\subsubsection{Dataset Composition and Splits}
To evaluate the models in a completely unseen scenario and assess their generalization capabilities, the \textit{Rooms} environment is used as a held-out test domain and therefore excluded from the training set.
The remaining data are split into training, validation, and test sets according to an 85\%–5\%–10\% ratio.
The resulting dataset, which we name PDDL2PRM, contains approximately one million reasoning steps, each annotated with a reward, providing a comprehensive resource for training and evaluating PRMs.

\subsubsection{Statistics}
\begin{table}[t]
\small
\centering
\begin{tabular}{l | c | c | c}

\toprule
\textbf{Domain} & \textbf{Problems} & \makecell{\textbf{MOPL}} & \textbf{Total steps}\\

\midrule

BlocksWorld-3 & ~~~~~952 & ~~5.29  & ~~51,990 \\
BlocksWorld-4 & ~~1,796 & ~~9.58 & 125,402 \\
Ferry & ~~~~~858 & ~~9.12 & ~~65,269 \\
Hanoi & ~~1,660 & ~~3.84 & ~~65,745 \\
Logistics & ~~~~~669 & ~~8.70 & ~~70,238 \\
Elevator & ~~2,089 & 10.53 & 207,549 \\
N-Puzzle & ~~~~~598 & ~~9.05 & ~~42,027 \\
Rooms & ~~~~~916 & ~~6.59 & ~~37,947 \\
Sokoban & ~~~~~437 & 14.25 & ~~58,475 \\
Spanner & ~~3,563 & ~~7.87 & 198,942 \\
VisitGrid & ~~1,176 & ~~5.86 & ~~61,390 \\

\midrule

\textbf{Total} & \textbf{14,714} & \textbf{~~7.94} & \textbf{984,974} \\

\bottomrule

\end{tabular}
\caption{\label{tab:domains_statistics}
Statistics for each domain, including the number of problems, the Mean Optimal Plan Length (MOPL), and the total number of alternatives (steps).
}
\end{table}
Because each planning domain exhibits intrinsic structural differences, both the number of generated problems and the length of their optimal solutions vary substantially. In many domains, increasing the number of objects, relations, or spatial degrees of freedom would in principle allow the generation of additional problems, but it would at the same time significantly increase the complexity of the resulting instances, often leading to much longer optimal plans. 
As a consequence, scaling the number of generated problems is not always practical, since we aim to avoid excessively long plans, which would produce reasoning trajectories that are difficult for models to handle.
This effect is particularly evident in domains such as \textit{Sokoban}, where the agent must navigate to precise grid locations before pushing boxes toward their targets. As the number of boxes increases, the sequence of required navigations and pushes grows accordingly, causing the optimal plan length to increase rapidly. Conversely, we observe that domains such as \textit{Spanner} and \textit{BlocksWorld} allow a broader variety of configurations without inducing the same degree of growth in plan length. The resulting statistics for each domain are reported in Table~\ref{tab:domains_statistics}.

\subsubsection{Computational Cost and Scalability of Optimal Planning}

From the perspective of computational cost and scalability, our approach offers several advantages.

First, dataset generation does not require GPU utilization for large-scale sampling from LLMs, which is often the dominant cost in alternative PRM supervision approaches. Instead, it is entirely CPU-bound, performed offline, and fully parallelizable across problem instances and domains.

Second, in the regime relevant for PRM training, optimal planning does not constitute a significant computational bottleneck. PDDL enables controlled domain design and systematic instance generation across varying difficulty levels, allowing us to regulate plan length and avoid instances that would be excessively costly for the planner.

We intentionally exclude domains and instances with extremely long optimal plans. This decision is not primarily driven by planning cost, but by reasoning considerations: long plans and large domains yield trajectories that are disproportionately difficult for PRMs to learn from. In practice, LLM reasoning degrades significantly as plan length increases~\cite{llm_reasoning_replace_planning, stechly2024cotplanning}.

More broadly, our goal is not to solve arbitrarily complex planning problems, but to generate large quantities of structured and logically coherent reasoning trajectories. This is achieved by sampling diverse yet tractable domains and instances.

Consequently, within the range of problem sizes relevant for PRM training, the effective complexity limit is dictated more by the reasoning capacity of PRMs than by the computational limits of the planner.

\subsection{Training Setup}
To increase the robustness and reliability of our results, we conduct experiments training two different models: \textit{Qwen2.5-Math-7B-Instruct}~\cite{Qwen2.5-Math-TR, qwen25math7binstruct} and \textit{Llama-3.1-8B-Instruct}~\cite{Llama-8B-Instruct, Llama-8B-Instruct-hf}, whose sizes align with the scale of most PRMs reported in the literature~\cite{Towards_Reasoning_Era}.

To evaluate the effectiveness of PDDL2PRM, we first train both models on a commonly adopted dataset (PRM800k or Math-Shepherd) as a point of comparison, and then retrain them on an augmented corpus that combines the same dataset with our PDDL-derived data.
Importantly, since PRM800k includes mathematical problems derived from the MATH test set, and one of our evaluation benchmarks, ProcessBench~\cite{ProcessBench}, is based on the same test samples, we follow the filtering approach adopted in ProcessBench and remove all MATH test samples from PRM800k to prevent train–test contamination.

Our training strategy follows a head-based architecture inspired by \citet{Qwen2.5-PRM}, which we adopt due to the strong performance reported in their work. In this setup, a head is added to the decoder to predict a scalar reward for each reasoning step. Unlike their approach, which formulates the task as classification, our PDDL-derived dataset provides multiple graded reward levels, enabling training PRMs as regressors. For comparison, we also train models using a classification loss to assess the effectiveness of the two formulations.
Each trained model is identified by a naming convention that specifies the underlying model, the training dataset(s), and a suffix indicating the learning objective, with \texttt{-r} denoting regression and \texttt{-c} denoting classification.
Further implementation details are provided in Appendix~\ref{appendix:appendix_C}.

We choose not to train the models exclusively on PDDL2PRM. Although the automatically translated actions are semantically accurate, they follow translation templates and lack the fluency and naturalness of CoT reasoning. 
For this reason, we find PDDL-derived data most effective when combined with existing resources, as it provides precise reasoning signals while other datasets contribute more natural and expressive reasoning patterns. Empirical results for a model trained solely on PDDL-derived data are reported in Appendix~\ref{appendix:appendix_B}.

\begin{table*}[ht]
\centering
\resizebox{0.98\textwidth}{!}{
\begin{tabular}{l|ccc|ccc|ccc|ccc|c}

\toprule

\textbf{Model} & \multicolumn{3}{c|}{\textbf{GSM8K}} & \multicolumn{3}{c|}{\textbf{MATH}} & \multicolumn{3}{c|}{\textbf{OlympiadBench}} & \multicolumn{3}{c|}{\textbf{Omni-MATH}} & \textbf{Avg. F1}\\
 & Error & Correct & F1 & Error & Correct & F1 & Error & Correct & F1 & Error & Correct & F1 \\
 
\midrule

Skywork-PRM-7B & 61.8 & 82.9 & 70.8 & 43.8 & 62.2 & 53.6 & 17.9 & 31.9 & 22.9 & 14.0 & 41.9 & 21.0 & 42.1\\

\midrule

Qwen2.5-Math-PRM-7B &72.0 & \textbf{96.4} & 82.4 & 68.0 & \textbf{90.4} & \textbf{77.6} & 55.7 & 85.5 & 67.5 & 55.2 & 83.0 & 66.3 & 73.5 \\
$^\dagger$Qwen2.5-Math-PRM-7B-PDDL-r & \textbf{73.4} & 95.9 & \textbf{83.2} & \textbf{68.2} & 89.2 & 77.3 & \textbf{57.9} & 81.1 & \textbf{67.6} & \textbf{57.6} & 78.8 & \textbf{66.5} & \textbf{73.6} \\

\midrule

Qwen2.5-Math-7B-PRM800k & 53.1 & 95.3 & 68.2 & 48.0 & 90.1 & 62.6 & 35.7 & \textbf{87.3} & 50.7 & 29.8 & \textbf{86.1} & 44.3 & 56.5 \\
$^\dagger$Qwen2.5-Math-7B-PRM800k-r & 58.0 & 89.6 & 70.4 & 60.8 & 74.4 & 66.9 & 48.4 & 56.0 & 52.0 & 49.7 & 57.3 & 53.2 & 60.6 \\
$^\dagger$Qwen2.5-Math-7B-PRM800k-PDDL-r & 67.6 & 90.2 & 77.3 & 67.2 & 73.4 & 70.1 & 51.7 & 53.1 & 52.4 & 50.6 & 52.7 & 51.6 & 62.9 \\

\midrule

$^\dagger$Llama-3.1-8B-PRM800k-r & 45.9 & 91.2 & 61.1 & 46.0 & 58.1 & 51.3 & 43.7 & 31.3 & 36.5 & 42.6 & 35.3 & 38.6 & 46.9 \\
$^\dagger$Llama-3.1-8B-PRM800k-PDDL-r & 58.5 & 91.7 & 71.4 & 53.9 & 55.4 & 54.6 & 46.3 & 27.1 & 34.2 & 42.7 & 32.4 & 36.8 & 49.3 \\

\midrule

$^\dagger$Qwen2.5-Math-7B-MathShepherd-r & 51.7 & 96.9 & 67.4 & 20.0 & 96.8 & 33.2 & ~~7.3 & 95.9 & 13.5 & ~~4.0 & 96.7 & ~~7.6 & 30.4 \\
$^\dagger$Qwen2.5-Math-7B-MathShepherd-PDDL-r & 58.0 & 93.8 & 71.7 & 34.7 & 88.9 & 49.9 & 19.2 & 81.1 & 31.1 & 13.6 & 79.7 & 23.2 & 44.0 \\

\bottomrule

\end{tabular}
}
\caption{Results on ProcessBench. $^\dagger$ indicates PRMs we trained; best results in \textbf{bold}.}

\label{tab:ProcessBench_results_all}
\end{table*}


\begin{table*}[ht]
\centering
\resizebox{0.85\textwidth}{!}{
\begin{tabular}{l|ccc|cc|c}

\toprule

\textbf{Model} & \textbf{Simplicity} & \textbf{Soundness} & \textbf{Sensitivity} & \textbf{Positive F1} & \textbf{Negative F1} & \textbf{Avg. F1}\\
 
\midrule

Skywork-PRM-7B & \textbf{81.7} & 56.6 & \textbf{90.1} & 89.2 & 40.9 & 65.1 \\

\midrule

Qwen2.5-Math-PRM-7B & 52.1 & 71.0 & 75.5 & \textbf{91.5} & 39.4 & 65.5 \\
$^\dagger$Qwen2.5-Math-PRM-7B-PDDL-r & 55.3 & \textbf{72.3} & 60.2 & 90.9 & \textbf{44.2} & \textbf{67.5} \\

\midrule

$^\dagger$Qwen2.5-Math-7B-MathShepherd-r & 48.5 & 55.1 & 51.9 & 81.5 & 26.5 & 54.0\\ 

$^\dagger$Qwen2.5-Math-7B-MathShepherd-PDDL-r  & 52.3 & 63.8 & 56.5 & 86.8 & 35.3 & 61.0 \\

\bottomrule

\end{tabular}
}
\caption{Results on PRMBench. $^\dagger$ indicates PRMs we trained; best results in \textbf{bold}. The full results on this benchmark are reported in Appendix~\ref{appendix:appendix_E}.}

\label{tab:PRMBench_results_all}
\end{table*}


\begin{table*}[ht]
\centering
\resizebox{0.89\textwidth}{!}{
\begin{tabular}{l | c | c | c | c | c | c | c | c}

\toprule

\textbf{Model} & \textbf{Math} & \textbf{Biology} & \textbf{Physics} & \textbf{Medicine} & \textbf{Chemistry} & \textbf{Logic} & \textbf{Avg. F1} & \textbf{Avg. F1} \\ 
& \textbf{F1} & \textbf{F1} & \textbf{F1} & \textbf{F1} 
& \textbf{F1} & \textbf{F1} & & \textbf{(no math)}\\

\midrule

Skywork-PRM-7B & 21.0 & 27.1 & 23.1 & 10.2 & 27.1 & 13.1 & 20.3 & 20.1 \\

\midrule

Qwen2.5-Math-PRM-7B & 55.5 & 22.8 & 36.4 & 23.5 & 39.5 & 19.0 & 32.8 & 28.2 \\
$^\dagger$Qwen2.5-Math-PRM-7B-PDDL-r & \textbf{56.1} & 29.6 & 36.2 & 26.7 & \textbf{42.7} & 24.0 & 35.9 & 31.8 \\

\midrule

Qwen2.5-Math-7B-PRM800k & 47.7 & 17.7 & 28.7 & 18.8 & 33.2 & ~~8.8 & 25.8 & 21.4 \\
$^\dagger$Qwen2.5-Math-7B-PRM800k-r & 47.5 & 21.9 & 40.0 & 23.9 & 29.7 & 22.4 & 30.9 & 27.6 \\
$^\dagger$Qwen2.5-Math-7B-PRM800k-PDDL-r & 53.6 & 27.1 & \textbf{45.0} & 30.0 & 37.6 & 27.6 & 36.8 & 33.5 \\

\midrule
$^\dagger$Llama-3.1-8B-PRM800k-r & 39.3 & 25.0 & 30.7 & 24.0 & 27.3 & 22.4 & 28.1 & 25.9 \\
$^\dagger$Llama-3.1-8B-PRM800k-PDDL-r & 44.4 & \textbf{36.6} & 42.1 & \textbf{35.5} & 36.4 & \textbf{28.1} & \textbf{37.2} & \textbf{35.7} \\

\midrule
$^\dagger$Qwen2.5-Math-7B-MathShepherd-r & 25.8 & ~~2.8 & ~~8.3 & ~~2.1 & ~~9.4 & ~~0.6 & ~~8.2 & ~~4.6 \\
$^\dagger$Qwen2.5-Math-7B-MathShepherd-PDDL-r & 43.2 & 13.3 & 33.6 & ~~9.0 & 20.6 & 15.6 & 22.5 & 18.4 \\

\bottomrule

\end{tabular}
}
\caption{Results on MR-Ben. $^\dagger$ indicates PRMs we trained; best results in \textbf{bold}. The full version of this table, including error and correct accuracies, is reported in Appendix~\ref{appendix:appendix_F}.}
\label{tab:mrben-nodet}
\end{table*}

\subsection{Test Data}
For evaluation, we use three different benchmarks. 

\textbf{ProcessBench}~\cite{ProcessBench} measures the ability to identify erroneous steps in mathematical CoT reasoning; given a math problem and a step-by-step solution, models must either identify the earliest-occurring error or conclude that all steps are correct. 
To evaluate model performance, the benchmark reports accuracy separately for instances containing at least one error in the CoT and for those with a fully correct reasoning chain.
It then computes their harmonic mean as an F1 score, providing a balanced measure of the model’s sensitivity to reasoning errors versus its tendency to incorrectly flag correct reasoning. 

The second benchmark, \textbf{PRMBench}~\cite{PRMBench}, is designed to evaluate PRMs by comparing their rewards against gold-standard annotations, providing a comprehensive assessment through metrics such as simplicity, soundness, and sensitivity.
The final score differs slightly from the previous benchmark, as it is computed as the arithmetic mean of the F1 score on instances containing erroneous reasoning steps and the F1 score on instances with fully correct CoT. Since PRMBench is constructed from PRM800k, all models trained on it are excluded from this evaluation.

\textbf{MR-Ben}~\cite{MR-Ben} evaluates PRMs across multiple domains, including physics, science, logic, and medicine. In its full setting, the benchmark requires models not only to identify the earliest erroneous step in a CoT, but also to provide an explanation of the error and a corresponding correction. However, given the structural constraints of the PRMs considered in this study, which output only scalar rewards, these explanatory and corrective components cannot be evaluated. Consequently, we restrict the evaluation to the error-identification task and score models using the same F1 metric as that described for ProcessBench.

Finally, we perform an additional evaluation on the PDDL2PRM test set and on previously excluded samples from the \textit{Rooms} domain to assess the ability of the PRMs to judge reasoning chains derived from PDDL problems. For consistency with the previous benchmarks, the models are tasked with identifying the first erroneous step in the CoT. Performance is measured using the same F1 metric as in ProcessBench and MR-Ben.

\section{Results}

\begin{table*}[ht]
\centering
\resizebox{0.76\textwidth}{!}{
\begin{tabular}{l | c c c | c c c | c}

\toprule

\textbf{Model} & \multicolumn{3}{c|}{\textbf{PDDL test set}} & \multicolumn{3}{c|}{\textbf{Rooms test set}} & \textbf{Avg. F1} \\

& \textbf{Error} & \textbf{Correct} & \textbf{F1} 
& \textbf{Error} & \textbf{Correct} & \textbf{F1} & \\

\midrule

Skywork-PRM-7B & 12.9 & ~~~~0.0 & ~~0.0 & 14.7 & ~~~~0.0 & ~~0.0 & ~~0.0 \\

\midrule
Qwen2.5-Math-PRM-7B & 32.5 & ~~68.2 & 44.0 & ~~9.5 & \textbf{100.0} & 17.3 & 30.7\\
$^\dagger$Qwen2.5-Math-PRM-7B-PDDL-r & 99.1 & ~~86.0 & 92.1 & \textbf{83.1} & ~~98.9 & \textbf{90.3} & \textbf{91.2} \\
\midrule

Qwen2.5-Math-7B-PRM800k & ~~0.8 & \textbf{100.0} & ~~1.6 & ~~0.0 & \textbf{100.0} & ~~0.0 & ~~0.8 \\
$^\dagger$Qwen2.5-Math-7B-PRM800k-r & 39.7 & ~~87.5 & 54.6 & 14.8 & \textbf{100.0} & 25.7 & 40.2 \\
$^\dagger$Qwen2.5-Math-7B-PRM800k-PDDL-r & \textbf{99.2} & ~~83.7 & 90.8 & 75.4 & ~~86.8 & 80.7 & 85.8 \\

\midrule
$^\dagger$Llama-3.1-8B-PRM800k-r & 49.4 & ~~51.1 & 50.2 & 46.9 & ~~99.5 & 63.7 & 57.0 \\
$^\dagger$Llama-3.1-8B-PRM800k-PDDL-r & \textbf{99.2} & ~~90.1 & \textbf{94.5} & 75.7 & ~~82.9 & 79.1 & 86.8 \\

\midrule
$^\dagger$Qwen2.5-Math-7B-MathShepherd-r & 14.9 & ~~~~1.6 & ~~2.9 & 14.7 & ~~~~0.0 & ~~0.0 & ~~1.5\\
$^\dagger$Qwen2.5-Math-7B-MathShepherd-PDDL-r & 96.8 & ~~83.0 & 89.3 & 81.3 & ~~99.9 & 89.6 & 89.5 \\

\bottomrule
\end{tabular}
}
\caption{Results on the PDDL test set. $^\dagger$ indicates PRMs we trained; best results in \textbf{bold}.}
\label{tab:testset_results_full_local_avg}
\end{table*}


We compare against the following models, all fine-tuned using \textit{Qwen2.5-Math-7B-Instruct} and selected as the most competitive existing PRMs with comparable parameter sizes:
\begin{itemize}
    \item \textbf{Skywork-PRM-7B}~\cite{he_2024_16998085}: with no details disclosed about the training procedure.
    
    \item \textbf{Qwen2.5-Math-7B-PRM800k}~\cite{ProcessBench}: trained as a binary classifier on the PRM800k dataset, where neutral annotations are mapped to the positive class.
    As anticipated, we fine-tune the same model on the same dataset while preserving the original labels and adopting a regression loss; our fine-tuned model is distinguished by the \texttt{-r} suffix.
      
    \item \textbf{Qwen2.5-Math-PRM-7B}~\cite{Qwen2.5-PRM}: trained on a large synthetic, non-public mathematical dataset and currently the strongest-performing PRM at this model scale. For experimental purposes, we further fine-tune this model on our PDDL-derived data.
\end{itemize}

\paragraph{Mathematical Reasoning Performance.} 
On ProcessBench, as reported in Table~\ref{tab:ProcessBench_results_all}, PRMs fine-tuned on the combined dataset that includes PDDL-derived data achieve an average improvement of +13.4\% compared to their non-PDDL counterparts.
The gains are particularly pronounced for models trained on Math-Shepherd, where the improvement is consistently larger.
Similar trends are observed on PRMBench, as shown in Table~\ref{tab:PRMBench_results_all}, with an average performance increase of +8.0\%.
These results are particularly relevant as they indicate that training PRMs to evaluate planning-oriented reasoning confers transferable benefits to mathematical reasoning evaluation, despite the PDDL-based data not being intrinsically mathematical.
\paragraph{Generalization Beyond Mathematics.} 
Results on MR-Ben (Table~\ref{tab:mrben-nodet}) show that, across domains such as biology, physics, medicine, chemistry, and logic, the benefits of incorporating PDDL-derived data are even more pronounced. In these categories, PRMs trained on the combined corpus outperform their non-PDDL counterparts by a substantial margin, with the largest improvements again observed for the models fine-tuned on Math-Shepherd.
Moreover, compared to the strongest public PRMs of comparable size, our PDDL-based models achieve superior performance, further highlighting the effectiveness of PDDL-derived supervision.

\paragraph{PDDL-Based Reasoning Evaluation} 
Finally, when evaluated on the PDDL test set, the impact of PDDL2PRM integration becomes striking. PRMs trained on the augmented dataset achieve substantial F1 improvements, demonstrating a strong ability to assess planning-based reasoning. These gains also generalize to unseen scenarios: on the \textit{Rooms} test set, performance improvements remain similarly pronounced.
In contrast, other PRMs perform poorly on this logic-intensive task. Models not trained by us, as well as our Math-Shepherd-only variant, achieve substantially lower performance. Some such models exhibit a strong bias toward classifying most reasoning chains as correct, performing well on error-free samples but poorly on samples with errors. Others display the opposite behavior, predicting chains as incorrect while struggling to localize the first erroneous step (Table~\ref{tab:testset_results_full_local_avg}).

\paragraph{Impact on a Strong Pre-trained PRM}
Fine-tuning the PRM introduced by \citet{Qwen2.5-PRM} with PDDL-derived data yields notable performance improvements. On mathematical benchmarks (Tables~\ref{tab:ProcessBench_results_all} and~\ref{tab:PRMBench_results_all}), performance remains stable, with slight improvements despite the non-mathematical nature of PDDL2PRM. In contrast, on non-mathematical benchmarks (Tables~\ref{tab:mrben-nodet} and~\ref{tab:testset_results_full_local_avg}), the PRM achieves substantial gains, further demonstrating the effectiveness of the proposed dataset.

\paragraph{Qualitative Analysis}
\label{sec:qual-analysis}
In addition to the quantitative experiments, we conducted a qualitative analysis on a sample of 100 instances drawn from multiple benchmarks and models, with a view to better understanding how PDDL-derived data affects PRMs' ability to evaluate free-form natural language reasoning.

We observe that PRMs trained with PDDL-derived data are more effective at identifying the first incorrect step in a reasoning chain, showing greater sensitivity to early errors than those without PDDL supervision. This suggests a stronger ability to localize failures at their source, consistent with the quantitative results on the erroneous chains reported in Tables \ref{tab:ProcessBench_results_all}, \ref{tab:PRMBench_results_all_3nd},
\ref{tab:mrben_results_full_local_avg_2}, and
\ref{tab:mrben_results_full_local_avg_3}.

Importantly, our qualitative analysis suggests that this gain is not limited to detecting arithmetic slips or local logical inconsistencies. Instead, PDDL-trained PRMs more reliably flag errors that result from an incorrect problem formulation, such as invalid assumptions, inappropriate reductions, or unjustified constraints that alter the structure of the task. In other words, they are better at catching modeling errors rather than only execution errors.

We hypothesize that this effect arises naturally from the planning formulation: PDDL makes preconditions and state transitions explicit, and training PRMs to judge whether an action violates formal constraints may encourage a verification style that transfers to natural language reasoning. Illustrative examples are provided in Appendix~\ref{app:qualitative_examples}.

\paragraph{Regression vs. Classification}
Notably, across all benchmarks, our model \textit{Qwen2.5-Math-7B-PRM800k-r} consistently outperforms the one proposed by \citet{ProcessBench}, \textit{Qwen2.5-Math-7B-PRM800k}. Both models are fine-tuned from the same model and on the same dataset, differing only in the learning objective: regression versus classification.
These results highlight the effectiveness of training PRMs as regressors when the supervision signal involves more than two distinct labels, as is the case for the PRM800k dataset. Additional experimental evidence is provided in Appendix~\ref{appendix:appendix_D}.

\section{Conclusion}
This work introduced a novel approach to PRM dataset construction leveraging PDDL domains as structured sources of reasoning data. We address key limitations of existing datasets, namely limited scalability due to high computational costs, narrow domain coverage, low granularity and precision.
We propose a synthetic data generation framework that produces reliable step-level rewards beyond the mathematical domain and allows the resulting dataset to be easily expanded through newly created PDDL domains and problems.
Evaluation across multiple benchmarks demonstrates that PRMs fine-tuned on a corpus augmented with PDDL-derived data consistently outperform their counterparts trained on the same starting data excluding the PDDL component. These results show that the proposed method and dataset effectively enhance the ability of PRMs to evaluate reasoning processes, highlighting PDDL-derived supervision as a reliable and promising resource for improving reasoning capabilities in LLMs.

\section*{Limitations} 
While our proposed approach provides several advantages, some limitations remain. First, the reward distribution is imbalanced: most examples receive reward values of 0.0, 0.5, or 1.0, while fewer instances fall into intermediate levels, such as 0.25 or 0.75. The 0.25 reward is intrinsically underrepresented, as several of the PDDL domains used are designed to remain solvable from almost any intermediate state, leaving very few actions that transition the system into an unsolvable configuration.

Second, the synthetic reasoning traces obtained by translating PDDL problems and actions into natural language lack the naturalness and fluency of CoTs produced by LLMs. Improving the linguistic quality of these translations is an important direction for future work, as it may enhance the alignment between synthetic supervision and the reasoning style of modern  LLMs.

Finally, another promising direction involves the automatic translation of natural-language problems into PDDL domains. Such a method would enable the dynamic generation of structured reasoning tasks tailored to specific problem types, further expanding the applicability and effectiveness of PDDL-based supervision.

\section*{Acknowledgments}
We gratefully acknowledge CINECA for providing the GPU computational resources, in particular through access to the HPC Leonardo infrastructure, which were essential for the development of this project.

\bibliography{custom}

\clearpage

\appendix

\section{Detailed Domain Descriptions}
\label{appendix:app_A}
This appendix provides detailed descriptions of the PDDL domains used to generate our dataset.

\subsection{BlocksWorld-4-ops}

The BlocksWorld-4-ops domain models a classical block-stacking scenario in which a robotic arm manipulates a set of blocks arranged on a table. Each block can either rest on the table or on top of another block, and at most one block may be held by the arm at any given time.

\paragraph{State Representation}
The state of the environment is described using the following predicates:
\begin{itemize}
    \item \texttt{on(x,y)}: block \texttt{x} is on top of block \texttt{y};
    \item \texttt{on-table(x)}: block \texttt{x} is on the table;
    \item \texttt{clear(x)}: block \texttt{x} has no block on top of it;
    \item \texttt{holding(x)}: the arm is holding block \texttt{x};
    \item \texttt{arm-empty}: the arm is not holding any block.
\end{itemize}

\paragraph{Actions}
The available actions are:
\begin{itemize}
    \item \texttt{pick-up(x)}: the arm picks up block \texttt{x} from the table; 
    \item \texttt{put-down(x)}: the arm places block \texttt{x} back on the table; 
    \item \texttt{stack(x,y)}: the arm places block \texttt{x} on top of block \texttt{y}; 
    \item \texttt{unstack(x,y)}: the arm removes block \texttt{x} from block \texttt{y}. 
\end{itemize}

\paragraph{Problem Instances}
Each problem instance defines an initial arrangement of blocks and a target configuration that the agent must reproduce through a sequence of manipulation actions.

\subsection{BlocksWorld-3-ops}

The BlocksWorld-3-ops domain represents a simplified variant of the block-stacking scenario introduced in the previous section. Unlike its 4-operator counterpart, this domain does not model an explicit robotic arm: blocks are moved directly between locations through abstract manipulation actions.

\paragraph{State Representation}
Since no arm is present, the state is encoded using a reduced set of predicates:
\begin{itemize}
    \item \texttt{on(x,y)}: block \texttt{x} is on top of block \texttt{y};
    \item \texttt{on-table(x)}: block \texttt{x} is on the table;
    \item \texttt{clear(x)}: block \texttt{x} has no block on top of it.
\end{itemize}

\paragraph{Actions}
The available actions are:
\begin{itemize}
    \item \texttt{move-b-to-t(x,y)}: block \texttt{x} is moved from on top of block \texttt{y} to the table; 
    
    \item \texttt{move-t-to-b(x,y)}: block \texttt{x} is moved from the table onto block \texttt{y}; 
    
    \item \texttt{move-b-to-b(x,y,z)}: block \texttt{x} is moved from on top of block \texttt{y} to on top of block \texttt{z}. 
\end{itemize}

\paragraph{Problem Instances}
As in the 4-operator version, each problem instance defines an initial configuration of blocks and a target arrangement that must be reconstructed, but here the plan consists solely of abstract move operations rather than fine-grained arm manipulations.

\subsection{Ferry}

The Ferry domain models a transportation scenario in which a set of cars must be moved between different locations using a single ferry. The ferry can carry at most one car at a time, and its position determines where boarding and disembarking operations may occur.

\paragraph{State Representation}
The state is characterized by predicates that track the locations of both the ferry and the cars:
\begin{itemize}
    \item \texttt{at-ferry(l)}: the ferry is currently at location \texttt{l};
    \item \texttt{at(c,l)}: car \texttt{c} is at location \texttt{l};
    \item \texttt{on(c)}: car \texttt{c} is loaded onto the ferry;
    \item \texttt{empty-ferry}: the ferry is not carrying a car.
\end{itemize}

\paragraph{Actions}
The available actions are:
\begin{itemize}
    \item \texttt{sail(l1,l2)}: moves the ferry from location \texttt{l1} to location \texttt{l2}; 
    \item \texttt{board(c,l)}: loads car \texttt{c} onto the ferry at location \texttt{l}; 
    \item \texttt{debark(c,l)}: unloads car \texttt{c} from the ferry at location \texttt{l}. 
\end{itemize}

\paragraph{Problem Instances}
Each problem instance specifies the initial locations of the cars and the ferry, together with a target configuration requiring specific cars to be delivered to designated locations.

\subsection{Tower of Hanoi}
The Tower of Hanoi domain models the classical puzzle in which disks of different sizes are stacked on pegs and must be moved to reach a target configuration while preserving the size ordering constraint.

\paragraph{State Representation}
The state of the environment is described using the following predicates:
\begin{itemize}
    \item \texttt{clear(x)}: no disk is on top of \texttt{x};
    \item \texttt{on(x,y)}: disk \texttt{x} is on top of \texttt{y} (which can be either another disk or a peg);
    \item \texttt{smaller(x,y)}: disk \texttt{x} is smaller than \texttt{y}.
\end{itemize}

\paragraph{Actions}
The available action is:
\begin{itemize}
    \item \texttt{move(d,f,t)}: the disk \texttt{d} is moved from support \texttt{f} to support \texttt{t}. 
\end{itemize}

\paragraph{Problem Instances}
Each problem instance defines an initial stacking of disks across the available pegs and a target configuration.

\subsection{Logistics}
The Logistics domain models a transportation scenario in which packages, trucks, and airplanes must be routed across different locations to achieve a specified delivery configuration. The environment is organized into cities, each containing several locations, and only airports allow inter-city movement.

\paragraph{State Representation}
The state of the environment is described using the following predicates:
\begin{itemize}
    \item \texttt{at(x,l)}: object or vehicle \texttt{x} is at location \texttt{l};
    \item \texttt{in(x,y)}: object \texttt{x} is loaded inside vehicle \texttt{y};
    \item \texttt{in-city(l,c)}: location \texttt{l} belongs to city \texttt{c};
\end{itemize}

\paragraph{Actions}
The available actions are:
\begin{itemize}
    \item \texttt{load-truck(o,t,l)}: loads the object \texttt{o} into the truck \texttt{t} when both are at location \texttt{l};
    \item \texttt{unload-truck(o,t,l)}: unloads the object \texttt{o} from truck \texttt{t} at location \texttt{l};
    \item \texttt{load-airplane(o,a,l)}: loads the object \texttt{o} into the airplane \texttt{a} when both are at location \texttt{l};
    \item \texttt{unload-airplane(o,a,l)}: unloads the object \texttt{o} from the airplane \texttt{a} at location \texttt{l};
    \item \texttt{drive-truck(t,l1,l2,c)}: moves the truck \texttt{t} from location \texttt{l1} to location \texttt{l2} within the city \texttt{c};
    \item \texttt{fly-airplane(a,l1,l2)}: flies the airplane \texttt{a} from location \texttt{l1} to location \texttt{l2}.
\end{itemize}

\paragraph{Problem Instances}
Each problem instance specifies the initial locations of all packages and vehicles, together with a target configuration that requires delivering the packages to their designated destinations.

\subsection{Elevator}
The Elevator domain (also known as Miconic) models a passenger transportation scenario in which an elevator must move across floors to pick up and drop off passengers according to their origin and destination floors.

\paragraph{State Representation}
The state of the environment is described using the following predicates:
\begin{itemize}
    \item \texttt{origin(p,f)}: passenger \texttt{p} is initially located at floor \texttt{f};
    \item \texttt{destin(p,f)}: floor \texttt{f} is the destination of passenger \texttt{p};
    \item \texttt{above(f1,f2)}: floor \texttt{f1} is above floor \texttt{f2};
    \item \texttt{boarded(p)}: passenger \texttt{p} is inside the elevator;
    \item \texttt{served(p)}: passenger \texttt{p} has reached their destination;
    \item \texttt{lift-at(f)}: the elevator is at floor \texttt{f}.
\end{itemize}

\paragraph{Actions}
The available actions are:
\begin{itemize}
    \item \texttt{board(f,p)}: passenger \texttt{p} boards the elevator at floor \texttt{f}; 
    \item \texttt{depart(f,p)}: passenger \texttt{p} exits the elevator at floor \texttt{f}; 
    \item \texttt{up(f1,f2)}: the elevator moves upward from floor \texttt{f1} to floor \texttt{f2}; 
    \item \texttt{down(f1,f2)}: the elevator moves downward from floor \texttt{f1} to floor \texttt{f2}; 
\end{itemize}

\paragraph{Problem Instances}
Each problem instance specifies the origin and destination floors of all passengers, the initial position of the elevator, and requires transporting every passenger to their assigned destination.

\subsection{N-Puzzle}
The N-Puzzle domain models a sliding-tile puzzle in which numbered tiles must be rearranged on a grid to reach a target configuration, using a single empty position to move tiles.

\paragraph{State Representation}
The state of the environment is described using the following predicates:
\begin{itemize}
    \item \texttt{at(t,p)}: tile \texttt{t} is at position \texttt{p};
    \item \texttt{empty(p)}: position \texttt{p} is currently empty.
\end{itemize}

\paragraph{Actions}
The available action is:
\begin{itemize}
    \item \texttt{move(t,p1,p2)}: tile \texttt{t} is moved from position \texttt{p1} to position \texttt{p2}. 
\end{itemize}

\paragraph{Problem Instances}
Each problem instance defines an initial placement of all tiles and the empty position on the grid, together with a target configuration that must be reached.

\subsection{VisitGrid}
The VisitGrid domain models a navigation task in which a robot must traverse a grid-like environment and visit a predefined set of locations.

\paragraph{State Representation}
The state of the environment is described using the following predicates:
\begin{itemize}
    \item \texttt{at-robot(l)}: the robot is currently located at location \texttt{l};
    \item \texttt{visited(l)}: location \texttt{l} has been visited.
\end{itemize}

\paragraph{Actions}
The available action is:
\begin{itemize}
    \item \texttt{move(l1,l2)}: moves the robot from location \texttt{l1} to location \texttt{l2}.
\end{itemize}

\paragraph{Problem Instances}
Each problem instance specifies the robot’s initial position and a set of target locations that must be visited at least once.

\subsection{Sokoban}
The Sokoban domain models a grid-based puzzle in which a robot must navigate through an environment with boxes and obstacles, pushing the boxes to new locations to reach a desired configuration.

\paragraph{State Representation}
The state of the environment is described using the following predicates:
\begin{itemize}
    \item \texttt{at-robot(l)}: the robot is at location \texttt{l};
    \item \texttt{at(b,l)}: box \texttt{b} is at location \texttt{l};
    \item \texttt{clear(l)}: location \texttt{l} does not contain a box.
\end{itemize}

\paragraph{Actions}
The available actions are:
\begin{itemize}
    \item \texttt{move(l1,l2,dir)}: the robot moves from location \texttt{l1} to location \texttt{l2};
    \item \texttt{push(l1,l2,b)}: the robot pushes box \texttt{b} from location \texttt{l1} to location \texttt{l2}.
\end{itemize}

\paragraph{Problem Instances}
Each problem instance specifies the initial positions of the robot and all boxes, together with a target configuration that requires placing each box in a designated location.

\subsection{Spanner}
The Spanner domain models a maintenance task in which an agent must move across a set of locations, collect spanners with limited durability, and use them to tighten a set of loose nuts distributed throughout the environment.

\paragraph{State Representation}
The state of the environment is described using the following predicates:
\begin{itemize}
    \item \texttt{at(x,l)}: object or agent \texttt{x} is at location \texttt{l};
    \item \texttt{loose(n)}: nut \texttt{n} is loose;
    \item \texttt{tightened(n)}: nut \texttt{n} has been tightened;
    \item \texttt{useable1(s)}, \texttt{useable2(s)}: spanner \texttt{s} has one or two remaining uses;
    \item \texttt{link(l1,l2)}: location \texttt{l1} is connected to location \texttt{l2}.
\end{itemize}

\paragraph{Actions}
The available actions are:
\begin{itemize}
    \item \texttt{move(actor,l1,l2)}: the agent moves from location \texttt{l1} to location \texttt{l2};
    \item \texttt{pick-up(actor,s)}: the agent picks up spanner \texttt{s};
    \item \texttt{tighten(actor,n,s)}: the agent tightens nut \texttt{n} using spanner \texttt{s}, consuming one unit of its durability.
\end{itemize}

\paragraph{Problem Instances}
Each problem instance defines the initial location of the agent, the placement and durability of the available spanners, and the set of loose nuts that must be tightened.

\subsection{Rooms}
The Rooms domain models a navigation task in which an agent moves across a collection of rooms connected by fragile doors and must turn off all lights in the environment before losing access to certain areas.

\paragraph{State Representation}
The state of the environment is described using the following predicates:
\begin{itemize}
    \item \texttt{at(a,r)}: the agent \texttt{a} is located in room \texttt{r};
    \item \texttt{door(r1,r2)}: rooms \texttt{r1} and \texttt{r2} are connected by a door;
    \item \texttt{door-intact(r1,r2)}: the door between \texttt{r1} and \texttt{r2} is intact and can still be traversed;
    \item \texttt{on(r)}: the light in room \texttt{r} is switched on.
\end{itemize}

\paragraph{Actions}
The available actions are:
\begin{itemize}
    \item \texttt{move(a,r1,r2)}: the agent moves from room \texttt{r1} to room \texttt{r2} breaking the traversal door;
    \item \texttt{turn-off(a,r)}: the agent turns off the light in room \texttt{r}.
\end{itemize}

\paragraph{Problem Instances}
Each problem instance specifies the initial position of the agent, the layout of the rooms and doors, and the subset of rooms whose lights are initially on and must be turned off.

\section{Training Implementation Details}
\label{appendix:appendix_C}
The architecture of the PRM was built using a head added on top of the decoder: the head maps the decoder’s final-layer hidden states to a scalar score for every token in the sequence. Concretely, it consists of two linear layers with an intermediate Tanh activation (Hidden $\rightarrow$ Hidden/2 $\rightarrow$ 1).
During the forward pass the computation proceeds as follows:
\begin{enumerate}
    \item The input sequence is fed to the base decoder, which returns last-layer hidden states.
    \item The head is applied token-wise to them, producing raw scalar logits (one per token).  
\end{enumerate}

The prompt, before being passed to the model, is enriched with special marker tokens that explicitly indicate the positions where a reward prediction is required (inserted immediately after each reasoning step, as done by \citet{OpenR}). Only these marker positions are relevant for the supervision: the trainer constructs a boolean mask that is 1 at marker positions and 0 elsewhere, and selects the corresponding predictions and targets.
Only predictions at marker positions contribute to the loss.
When trained in the regression setting, the model minimizes the Mean Squared Error (MSE) between predicted rewards and target scores. In the classification setting, we instead optimize the Binary Cross-Entropy (BCE) loss with logits.

Fine-tuning was conducted using Low-Rank Adaptation (LoRA) \cite{Lora}.
Both the LoRA adapters and the head weights are updated, while the remaining model parameters remain frozen.
The parameters used for LoRA are shown in Table \ref{tab:lora-config}.\\

\begin{table}[h!]
\centering
\begin{tabular}{|c|p{3cm}|}
\hline
\textbf{Parameter} & \textbf{Value} \\
\hline
Rank & 64 \\
LoRA scaling factor ($\alpha$) & 64 \\
Target modules & Q, K, V, O of the transformer + main feed-forward layers \\
LoRA dropout & 0.1 \\
\hline
\end{tabular}
\caption{LoRA configuration.}
\label{tab:lora-config}
\end{table}
All models were fine-tuned using FP16 precision.
The learning rate was selected in the range $[1 \times 10^{-5}, 3 \times 10^{-5}]$ and adjusted based on the size of the training dataset.
A batch size of 32 and a weight decay of 0.05 were used in all experiments.

The PRM800k dataset was pre-processed to match the structure of the PDDL-derived dataset and the Math-Shepherd dataset, including the removal of incomplete samples and entries lacking step-level reward annotations. 
Each resulting sample consists of a problem description and a (possibly partial) CoT comprising a variable number of reasoning steps, along with the corresponding step-level rewards. 
After filtering, the processed PRM800k dataset contains approximately 350k samples. 
On the other hand, the Math-Shepherd dataset comprises approximately 400k samples.
To ensure a balanced training process when combining PDDL2PRM with either PRM800k or Math-Shepherd, the PDDL-derived dataset was not used in its entirety. 
Instead, we randomly sampled a subset of instances from the PDDL2PRM training corpus whose size matches that of the paired dataset used during training.

\section{Ablation Study: PDDL-Only SFT}
\label{appendix:appendix_B}

In this appendix, we report results for a model trained exclusively on PDDL-derived supervision. 
The architecture, regression-based training setup, and optimization procedure are identical to those used in the main experiments. 
The only difference lies in the training data, which consists solely of the PDDL-derived dataset that is used as part of the mixed supervision for the other models.

As shown in Tables~\ref{tab:ProcessBench_results_pddlonly}, \ref{tab:mrben-pddlonly}, and~\ref{tab:testset_onlypddl}, the model trained exclusively on PDDL-derived data underperforms models trained with natural-language reasoning supervision, such as PRM800k or Math-Shepherd, across the benchmarks. This performance gap is particularly pronounced on the mathematical evaluations of ProcessBench, which fall outside the scope of PDDL-based supervision. An exception is MR-Ben, where the PDDL-trained model outperforms the Math-Shepherd–trained model, as well as the PDDL reasoning evaluation, on which the PDDL-trained model achieves superior performance.

\section{Regression vs. Classification Detailed Results}
\label{appendix:appendix_D}
To further investigate the differences between regression- and classification-based training, we additionally trained two classification-based PRMs. Specifically, we trained one model starting from \textit{Llama-3.1-8B-Instruct} on PRM800k by mapping neutral labels to the positive class, following the procedure adopted by \citet{ProcessBench}, and another model starting from \textit{Qwen2.5-Math-7B-Instruct} on Math-Shepherd to enable a direct comparison with the regression-trained PRMs presented earlier.

Tables~\ref{tab:ProcessBench_regvsclass}, \ref{tab:mrben_regvsclass}, and \ref{tab:PDDL_regvsclass} report a detailed comparison between regression- and classification-based PRM training.
For PRM800k, regression consistently outperforms classification across all benchmarks, as the dataset provides three distinct supervision levels that are more naturally captured by a regression objective.
In contrast, for Math-Shepherd the results are nearly identical, with a slight advantage for the classification setting; this outcome is expected, since the dataset inherently contains only two labels.
Overall, these results confirm that regression losses are more suitable when supervision involves multiple graded labels.

\section{PRMBench Detailed Results}
\label{appendix:appendix_E}
Tables~\ref{tab:PRMBench_results_all_2nd} and \ref{tab:PRMBench_results_all_3nd} provide a detailed breakdown of the results on the PRMBench benchmark.
In particular, \textit{Positive Acc} and \textit{Negative Acc} denote accuracy on positive and negative reasoning steps, respectively, while \textit{First} measures accuracy in identifying the first erroneous step in the CoT.
\textit{Similarity} is a measure of a model’s ability to distinguish between correct and incorrect reasoning steps.
\textit{Positive F1} and \textit{Negative F1} report the F1 scores on positive and negative steps, respectively.

In addition, PRMBench reports several diagnostic metrics:
\begin{itemize}
    \item \textbf{NR (Non-Redundancy):} evaluates the PRM’s ability to identify redundant steps within the reasoning process.
    \item \textbf{NCL (Non-Circular Logic):} assesses the PRM’s ability to detect circular reasoning.
    \item \textbf{ES (Empirical Soundness):} measures the ability to detect counterfactual errors within the reasoning process.
    \item \textbf{SC (Step Consistency):} evaluates whether the PRM can detect step-wise contradictions.
    \item \textbf{DC (Domain Consistency):} measures the ability to identify domain-inconsistent reasoning, a specific type of counterfactual error.
    \item \textbf{CI (Confidence Invariance):} evaluates whether the PRM can detect over-confident errors.
    \item \textbf{PS (Prerequisite Sensitivity):} measures sensitivity to missing conditions or prerequisite violations.
    \item \textbf{DR (Deception Resistance):} evaluates the ability to detect deceptive or misleading reasoning steps.
    \item \textbf{MS (Multi-Solution Consistency):} assesses consistency across different solution paths for the same problem.
\end{itemize}

\section{MR-Ben Detailed Results}
\label{appendix:appendix_F}
Tables~\ref{tab:mrben_results_full_local_avg_2} and \ref{tab:mrben_results_full_local_avg_3} report the complete results on MR-Ben, including accuracy on fully correct CoTs as well as accuracy on chains containing reasoning errors.

\section{Computational Costs}
\label{appendix:appendix_G}

We report the computational resources required to train the PRMs used in this work. All models were trained using A100 GPUs. For each training run, we used 8 GPUs in parallel.

Training time scales approximately linearly with the size of the training corpus. For models trained on the PRM800k dataset or on the Math-Shepherd dataset alone, training required between 4 and 6 hours, corresponding to approximately 32–48 GPU-hours per model. Models trained on the combined corpus required between 7 and 10 hours, corresponding to approximately 56–80 GPU-hours.

Training was performed using standard distributed data-parallel setups. Limited hyperparameter tuning was performed to adjust key training parameters, including the learning rate, weight decay, and LoRA dropout.

\section{Additional Qualitative Examples}
\label{app:qualitative_examples}

\begin{figure*}
    \centering
    \includegraphics[width=1\linewidth]{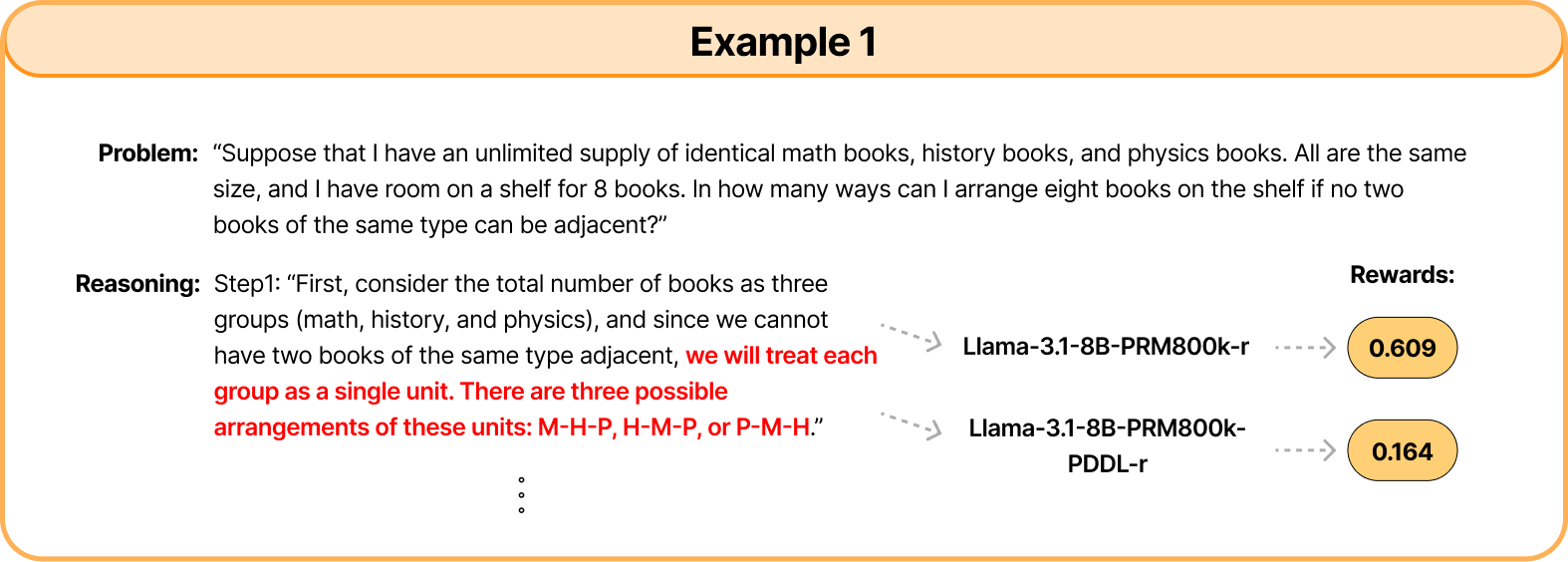}
    \caption{Example from the MATH subset of ProcessBench. We report scores from two PRMs based on Llama + PRM800k. The PDDL-trained PRM more strongly penalizes the incorrect combinatorial formulation, where the reasoning restricts the solution space to permutations of fixed groups, excluding valid interleavings across types.}
    \label{fig:ex1}
\end{figure*}

\begin{figure*}
    \centering
    \includegraphics[width=1\linewidth]{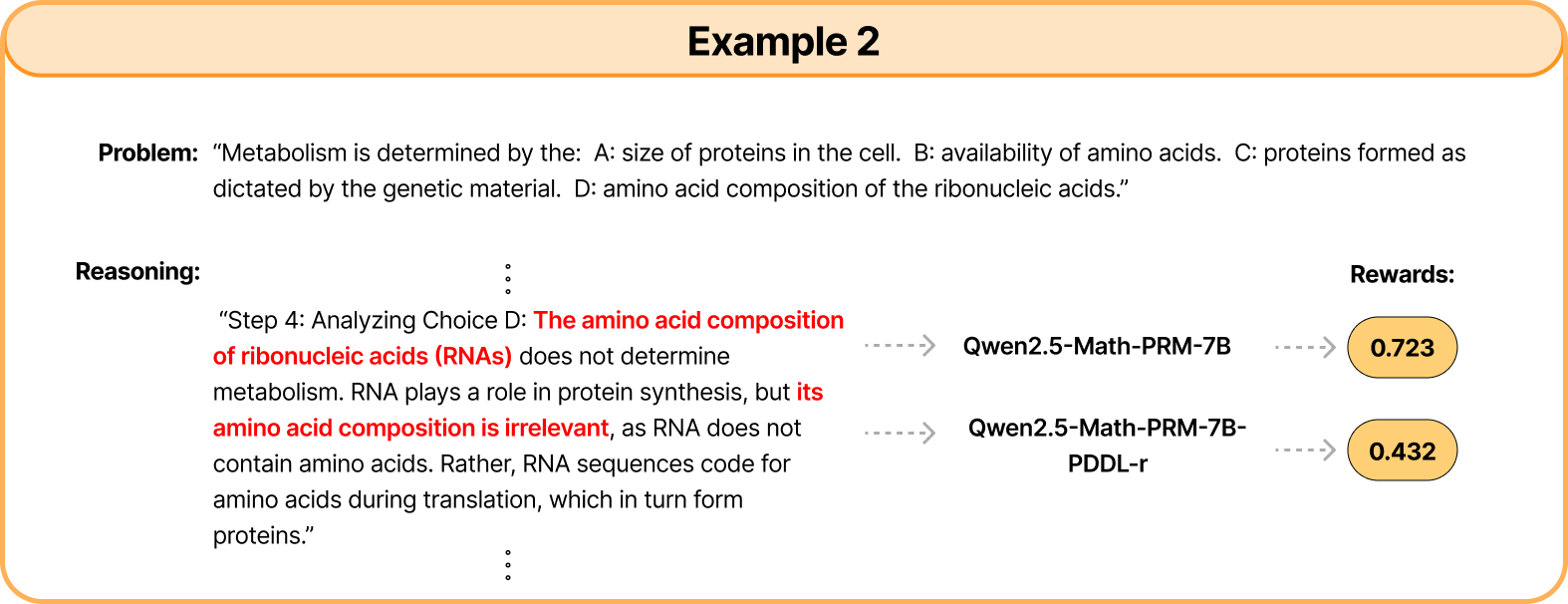}
    \caption{Example from the Medicine subset of MR-Ben. We report scores from two PRMs based on Qwen2.5-Math-PRM-7B. The PDDL-trained PRM assigns a lower score to the fourth reasoning step, correctly penalizing a biologically incorrect justification that conflates RNA composition with its functional role in protein synthesis.}
    \label{fig:ex2}
\end{figure*}

To further illustrate the qualitative pattern discussed in Section~\ref{sec:qual-analysis}, we report two representative examples in Figures~\ref{fig:ex1} and~\ref{fig:ex2}. In both cases, the PRM trained with PDDL-derived supervision more reliably penalizes reasoning steps that introduce an incorrect formulation of the problem itself. In Figure~\ref{fig:ex1}, the chain begins by imposing an invalid combinatorial structure on the task, collapsing the space of valid arrangements into three fixed groups and thereby excluding legitimate alternating patterns.

In Figure~\ref{fig:ex2}, the reasoning introduces a biologically invalid conceptual statement, which establishes incorrect foundations for the subsequent argument.

In both examples, this difference is reflected in the reward trajectories: the PRM trained without PDDL supervision continues to assign relatively high scores even after the incorrect formulation has been introduced, whereas the PRM trained with PDDL supervision drops more sharply once the chain departs from a valid problem representation.

\section{Licenses}
\label{appendix:appendix_H}
In this work we leverage several existing datasets and pretrained models. Below we summarize their licensing terms in accordance with the information provided by the original authors and public repositories:

\begin{itemize}
  \item \textbf{PRM800k}: Dataset released under the MIT License.

  \item \textbf{Math-Shepherd}: Publicly available dataset released without an explicit license. It is used strictly for research and experimental purposes.

  \item \textbf{Skywork-o1-Open-PRM-Qwen-2.5-7B}: Released under the Skywork Community License, that permits research and experimental use.

  \item \textbf{Qwen2.5-Math models}: Released under the Apache License, Version 2.0.

  \item \textbf{meta-llama/Llama-3.1-8B-Instruct}: Distributed under the Meta LLaMA 3.1 Community License, which permits research and experimental use.
\end{itemize}

The dataset and trained models introduced in this work will be released under the Creative Commons Attribution-ShareAlike (CC BY-SA 4.0) license.

\begin{table*}[ht]
\centering
\resizebox{\textwidth}{!}{
\begin{tabular}{l|ccc|ccc|ccc|ccc|c}

\toprule

\textbf{Model} & \multicolumn{3}{c|}{\textbf{GSM8K}} & \multicolumn{3}{c|}{\textbf{MATH}} & \multicolumn{3}{c|}{\textbf{OlympiadBench}} & \multicolumn{3}{c|}{\textbf{Omni-MATH}} & \textbf{Avg. F1}\\
 & Error & Correct & F1 & Error & Correct & F1 & Error & Correct & F1 & Error & Correct & F1 \\
 
\midrule

Qwen2.5-Math-7B-MathShepherd-r & 51.7 & 96.9 & 67.4 & 20.0 & 96.8 & 33.2 & ~~7.3 & 95.9 & 13.5 & ~~4.0 & 96.7 & ~~7.6 & 30.4 \\

Qwen2.5-Math-7B-PRM800k-r & 58.0 & 89.6 & 70.4 & 60.8 & 74.4 & 66.9 & 48.4 & 56.0 & 52.0 & 49.7 & 57.3 & 53.2 & 60.6 \\

Qwen2.5-Math-7B-PDDL-r & 33.3 & ~~9.3 & 14.6 & 24.4 & 22.7 & 23.5 & 15.4 & 35.1 & 21.4 & ~~9.2 & 28.6 & 14.0 & 18.4 \\

\bottomrule

\end{tabular}
}
\caption{Results of the PDDL-only model on ProcessBench compared with MathShepherd and PRM800k models.}
\label{tab:ProcessBench_results_pddlonly}
\end{table*}


\begin{table*}[ht]
\centering
\resizebox{0.9\textwidth}{!}{
\begin{tabular}{l | c | c | c | c | c | c | c | c}

\toprule

\textbf{Model} & \textbf{Math} & \textbf{Biology} & \textbf{Physics} & \textbf{Medicine} & \textbf{Chemistry} & \textbf{Logic} & \textbf{Avg. F1} & \textbf{Avg. F1} \\ 
& \textbf{F1} & \textbf{F1} & \textbf{F1} & \textbf{F1} 
& \textbf{F1} & \textbf{F1} & & \textbf{(no math)}\\

\midrule

Qwen2.5-Math-7B-MathShepherd-r & 25.8 & ~~2.8 & ~~8.3 & ~~2.1 & ~~9.4 & ~~0.6 & ~~8.2 & ~~4.6 \\
Qwen2.5-Math-7B-PRM800k-r & 47.5 & 21.9 & 40.0 & 23.9 & 29.7 & 22.4 & 30.9 & 27.6 \\
Qwen2.5-Math-7B-PDDL-r & 26.3 & ~~7.8 & 21.8 & 11.3 & 12.5 & 10.5 & 15.0 & 12.8 \\
\bottomrule

\end{tabular}
}
\caption{Results of the PDDL-only model with Math-Shepherd and PRM800k models on the MR-Ben benchmark.
}
\label{tab:mrben-pddlonly}
\end{table*}


\begin{table*}[ht]
\centering
\resizebox{0.85\textwidth}{!}{
\begin{tabular}{l | c c c | c c c | c}

\toprule

\textbf{Model} & \multicolumn{3}{c|}{\textbf{PDDL test set}} & \multicolumn{3}{c|}{\textbf{Rooms test set}} & \textbf{Avg. F1} \\

& \textbf{Error} & \textbf{Correct} & \textbf{F1} 
& \textbf{Error} & \textbf{Correct} & \textbf{F1} & \\

\midrule

Qwen2.5-Math-7B-MathShepherd-r & 14.9 & ~~1.6 & ~~2.9 & 14.7 & ~~~~0.0 & ~~0.0 & ~~1.5\\

Qwen2.5-Math-7B-PRM800k-r & 39.7 & 87.5 & 54.6 & 14.8 & 100.0 & 25.7 & 40.2 \\

Qwen2.5-Math-7B-PDDL-r & 98.9 & 84.9 & 91.3 & 78.3 & ~~90.6 & 84.0 & 87.7 \\

\bottomrule
\end{tabular}
}
\caption{Performance of the PDDL-only model on the PDDL and Rooms test sets, compared with models trained on Math-Shepherd and PRM800k.
}
\label{tab:testset_onlypddl}
\end{table*}


\begin{table*}[ht]
\centering
\resizebox{\textwidth}{!}{
\begin{tabular}{l|ccc|ccc|ccc|ccc|c}

\toprule

\textbf{Model} & \multicolumn{3}{c|}{\textbf{GSM8K}} & \multicolumn{3}{c|}{\textbf{MATH}} & \multicolumn{3}{c|}{\textbf{OlympiadBench}} & \multicolumn{3}{c|}{\textbf{Omni-MATH}} & \textbf{Avg. F1}\\
 & Error & Correct & F1 & Error & Correct & F1 & Error & Correct & F1 & Error & Correct & F1 \\
 
\midrule

Qwen2.5-Math-7B-PRM800k & 53.1 & 95.3 & 68.2 & 48.0 & 90.1 & 62.6 & 35.7 & 87.3 & 50.7 & 29.8 & 86.1 & 44.3 & 56.5 \\
$^\dagger$Qwen2.5-Math-7B-PRM800k-r & \textbf{58.0} & 89.6 & \textbf{70.4} & \textbf{60.8} & 74.4 & \textbf{66.9} & \textbf{48.4} & 56.0 & \textbf{52.0} & \textbf{49.7} & 57.3 & \textbf{53.2} & \textbf{60.6} \\

\midrule
$^\dagger$Llama-3.1-8B-PRM800k-c & 39.1 & 89.1 & 54.4 & 43.4 & 58.1 & 49.7 & 40.7 & 35.4 & 37.9 & 39.9 & 38.6 & 39.2 & 45.3 \\

$^\dagger$Llama-3.1-8B-PRM800k-r & 45.9 & 91.2 & 61.1 & 46.0 & 58.1 & 51.3 & 43.7 & 31.3 & 36.5 & 42.6 & 35.3 & 38.6 & 46.9 \\

\midrule
$^\dagger$Qwen2.5-Math-7B-MathShepherd-c & 50.2 & \textbf{97.4} & 66.3 & 19.9 & 96.6 & 33.0 & ~~7.4 & \textbf{96.2} & 13.8 & ~~4.9 & 95.9 & ~~9.3 & 30.6 \\

$^\dagger$Qwen2.5-Math-7B-MathShepherd-r & 51.7 & 96.9 & 67.4 & 20.0 & \textbf{96.8} & 33.2 & ~~7.3 & 95.9 & 13.5 & ~~4.0 & \textbf{96.7} & ~~7.6 & 30.4 \\

\bottomrule

\end{tabular}
}
\caption{Classification vs. Regression on ProcessBench. $^\dagger$ indicates PRMs we trained; best results in \textbf{bold}. (Qwen2.5-Math-7B-PRM800k is trained as a classifier).}

\label{tab:ProcessBench_regvsclass}
\end{table*}


\begin{table*}[ht]
\centering
\resizebox{0.95\textwidth}{!}{
\begin{tabular}{l | c | c | c | c | c | c | c | c}

\toprule

\textbf{Model} & \textbf{Math} & \textbf{Biology} & \textbf{Physics} & \textbf{Medicine} & \textbf{Chemistry} & \textbf{Logic} & \textbf{Avg. F1} & \textbf{Avg. F1} \\ 
& \textbf{F1} & \textbf{F1} & \textbf{F1} & \textbf{F1} 
& \textbf{F1} & \textbf{F1} & & \textbf{(no math)}\\

\midrule

Qwen2.5-Math-7B-PRM800k & \textbf{47.7} & 17.7 & 28.7 & 18.8 & \textbf{33.2} & ~~8.8 & 25.8 & 21.4 \\
$^\dagger$Qwen2.5-Math-7B-PRM800k-r & 47.5 & 21.9 & \textbf{40.0} & 23.9 & 29.7 & \textbf{22.4} & \textbf{30.9} & \textbf{27.6} \\

\midrule

$^\dagger$Llama-3.1-8B-PRM800k-c & 35.0 & 20.6 & 33.4 & \textbf{25.8} & 26.4 & 20.1 & 26.9 & 25.3 \\
$^\dagger$Llama-3.1-8B-PRM800k-r & 39.3 & \textbf{25.0} & 30.7 & 24.0 & 27.3 & \textbf{22.4} & 28.1 & 25.9 \\

\midrule
$^\dagger$Qwen2.5-Math-7B-MathShepherd-c & 25.9 & ~~2.8 & ~~7.7 & ~~1.7 & ~~9.8 & ~~1.4 & ~~8.2 & ~~4.7 \\
$^\dagger$Qwen2.5-Math-7B-MathShepherd-r & 25.8 & ~~2.8 & ~~8.3 & ~~2.1 & ~~9.4 & ~~0.6 & ~~8.2 & ~~4.6 \\

\bottomrule

\end{tabular}
}
\caption{Classification vs. Regression on MR-Ben. $^\dagger$ indicates PRMs we trained; best results in \textbf{bold}. (Qwen2.5-Math-7B-PRM800k is trained as a classifier).}
\label{tab:mrben_regvsclass}
\end{table*}


\begin{table*}[ht]
\centering
\resizebox{0.82\textwidth}{!}{
\begin{tabular}{l | c c c | c c c | c}

\toprule

\textbf{Model} & \multicolumn{3}{c|}{\textbf{PDDL test set}} & \multicolumn{3}{c|}{\textbf{Rooms test set}} & \textbf{Avg. F1} \\

& \textbf{Error} & \textbf{Correct} & \textbf{F1} 
& \textbf{Error} & \textbf{Correct} & \textbf{F1} & \\

\midrule

Qwen2.5-Math-7B-PRM800k & ~~0.8 & \textbf{100.0} & ~~1.6 & ~~0.0 & \textbf{100.0} & ~~0.0 & ~~0.8 \\
$^\dagger$Qwen2.5-Math-7B-PRM800k-r & 39.7 & ~~87.5 & \textbf{54.6} & 14.8 & \textbf{100.0} & 25.7 & 40.2 \\

\midrule

$^\dagger$Llama-3.1-8B-PRM800k-c & \textbf{51.0} & ~~34.8 & 41.4 & 45.4 & ~~99.4 & 62.3 & 51.9 \\
$^\dagger$Llama-3.1-8B-PRM800k-r & 49.4 & ~~51.1 & 50.2 & \textbf{46.9} & ~~99.5 & \textbf{63.7} & \textbf{57.0} \\

\midrule

$^\dagger$Qwen2.5-Math-7B-MathShepherd-c & 14.8 & ~~~~1.0 & ~~1.9 & 14.8 & ~~~~0.1 & ~~0.1 & ~~1.0 \\
$^\dagger$Qwen2.5-Math-7B-MathShepherd-r & 14.9 & ~~~~1.6 & ~~2.9 & 14.7 & ~~~~0.0 & ~~0.0 & ~~1.5\\

\bottomrule
\end{tabular}
}
\caption{Classification vs. Regression on PDDL test set. $^\dagger$ indicates PRMs we trained; best results in \textbf{bold}. (Qwen2.5-Math-7B-PRM800k is trained as a classifier).}
\label{tab:PDDL_regvsclass}
\end{table*}


\begin{table*}[ht]
\centering
\resizebox{0.98\textwidth}{!}{
\begin{tabular}{l|ccc|ccccc|cccc|c}

\toprule

\textbf{Model} & \multicolumn{3}{c|}{\textbf{Simplicity}} & \multicolumn{5}{c|}{\textbf{Soundness}} & \multicolumn{4}{c|}{\textbf{Sensitivity}} & \textbf{Overall} \\
& NR. & NCL. & Avg. & ES & SC. & DC. & CI & Avg. & PS & DR. & MS. & Avg. & \\
 
\midrule

Skywork-PRM-7B & 56.4 & \textbf{62.8} & 59.6 & 69.4 & 67.1 & 67.7 & 69.9 & 68.5 & \textbf{60.9} & 65.8 & 93.2 & 73.3 & 65.1 \\

\midrule

$^\dagger$Qwen2.5-Math-7B-PDDL-r & \textbf{59.1} & 60.8 & \textbf{59.9} & 63.3 & 62.0 & 62.5 & 65.9 & 63.4 & 59.2 & 58.7 & 42.5 & 53.5 & 61.4 \\

\midrule

Qwen2.5-Math-PRM-7B & 49.0 & 55.1 & 52.1 & 71.8 & 67.3 & 66.3 & 78.5 & 71.0 & 57.6 & 69.1 & \textbf{99.7} & \textbf{75.5} & 65.5 \\
$^\dagger$Qwen2.5-Math-PRM-7B-PDDL-r & 50.6 & 60.1 & 55.3 & \textbf{73.7} & \textbf{69.1} & \textbf{67.8} & \textbf{78.6} & \textbf{72.3} & 60.2 & \textbf{70.5} & 49.9 & 60.2 & \textbf{67.5}\\

\midrule

$^\dagger$Qwen2.5-Math-7B-MathShepherd-c & 47.4 & 49.5 & 48.5 & 56.3 & 51.6 & 51.3 & 57.5 & 54.2 & 51.9 & 56.3 & 46.9 & 51.7 & 53.4 \\

$^\dagger$Qwen2.5-Math-7B-MathShepherd-r & 46.7 & 50.2 & 48.5 & 57.5 & 52.6 & 51.4 & 59.0 & 55.1 & 51.7 & 56.8 & 47.2 & 51.9 & 54.0 \\ 

$^\dagger$Qwen2.5-Math-7B-MathShepherd-PDDL-r & 48.8 & 55.7 & 52.3 & 66.7 & 60.7 & 60.4 & 67.7 & 63.8 & 56.8 & 62.9 & 49.7 & 56.5 & 61.0 \\

\bottomrule

\end{tabular}
}
\caption{Simplicity, Soundness and Sensitivity results on PRMBench. $^\dagger$ indicates PRMs we trained; best results in \textbf{bold}.}

\label{tab:PRMBench_results_all_2nd}
\end{table*}

\begin{table*}[ht]
\centering
\resizebox{\textwidth}{!}{
\begin{tabular}{l|c|c|c|c|c|c|c|c}

\toprule

\textbf{Model} & \textbf{Total Acc} & \textbf{Positive Acc} & \textbf{Negative Acc} & \textbf{First} & \textbf{Similarity} & \textbf{Positive F1} & \textbf{Negative F1} & \textbf{PRMScore}\\
 
\midrule

Skywork-PRM-7B & 81.7 & 88.5 & 42.7 & \textbf{56.6} & 90.1 & 89.2 & 40.9 & 65.1 \\

\midrule

$^\dagger$Qwen2.5-Math-7B-PDDL-r & 75.9 & 81.4 & \textbf{46.2} & 53.3 & 84.6 & 85.0 & 37.7 & 61.4 \\

\midrule

Qwen2.5-Math-PRM-7B & \textbf{85.1} & \textbf{95.4} & 30.6 & 37.8 & 89.0 & \textbf{91.5} & 39.4 & 65.5 \\
$^\dagger$Qwen2.5-Math-PRM-7B-PDDL-r & 84.4 & 92.8 & 39.2 & 47.4 & 86.3 & 90.9 & \textbf{44.2} & \textbf{67.5} \\

\midrule
$^\dagger$Qwen2.5-Math-7B-MathShepherd-c & 68.6 & 74.6 & 36.5 & 33.1 & 84.4 & 80.0 & 26.8 & 53.4 \\

$^\dagger$Qwen2.5-Math-7B-MathShepherd-r & 70.4 & 77.3 & 33.9 & 30.5 & 84.6 & 81.5 & 26.5 & 54.0 \\ 

$^\dagger$Qwen2.5-Math-7B-MathShepherd-PDDL-r & 78.0 & 85.6 & 37.9 & 38.9 & \textbf{84.0} & 86.8 & 35.3 & 61.0 \\

\bottomrule

\end{tabular}
}
\caption{Detailed results on PRMBench. $^\dagger$ indicates PRMs we trained; best results in \textbf{bold}.}

\label{tab:PRMBench_results_all_3nd}
\end{table*}

\begin{table*}[ht]
\centering
\resizebox{0.9\textwidth}{!}{
\begin{tabular}{l | c c c | c c c | c c c }

\toprule

Model & \multicolumn{3}{c|}{\textbf{Math}} 
& \multicolumn{3}{c|}{\textbf{Biology}}
& \multicolumn{3}{c}{\textbf{Physics}} 
\\
& \textbf{Error} & \textbf{Correct} & \textbf{F1} 
& \textbf{Error} & \textbf{Correct} & \textbf{F1} 
& \textbf{Error} & \textbf{Correct} & \textbf{F1}
\\

\midrule

Skywork-PRM-7B & 28.1 & 16.8 & 21.0 & \textbf{26.6} & 27.7 & 27.1 & 23.0 & 23.2 & 23.1 \\

\midrule

Qwen2.5-Math-PRM-7B & 45.5 & 71.1 & 55.5 & 13.2 & 86.1 & 22.8 & 24.4 & 71.8 & 36.4 \\
$^\dagger$Qwen2.5-Math-PRM-7B-PDDL-r & \textbf{48.9} & 65.8 & \textbf{56.1} & 18.4 & 75.7 & \textbf{29.6} & 25.3 & 63.6 & 36.2 \\

\midrule

Qwen2.5-Math-7B-PRM800k & 36.8 & 67.8 & 47.7 & ~~9.8 & 90.3 & 17.7 & 18.1 & 69.3 & 28.7 \\
$^\dagger$Qwen2.5-Math-7B-PRM800k-r & 37.8 & 63.8 & 47.5 & 13.6 & 56.2 & 21.9 & 30.7 & 57.4 & 40.0 \\
$^\dagger$Qwen2.5-Math-7B-PRM800k-PDDL-r & 47.3 & 61.7 & 53.6 & 17.7 & 57.7 & 27.1 & \textbf{39.7} & 52.0 & \textbf{45.0} \\ 
$^\dagger$Qwen2.5-Math-7B-PDDL-r & 20.7 & 36.2 & 26.3 & ~~4.1 & 83.3 & ~~7.8 & 13.2 & 61.8 & 21.8 \\

\midrule

$^\dagger$Llama-3.1-8B-PRM800k-c & 23.5 & 68.5 & 35.0 & 11.7 & 85.7 & 20.6 & 22.4 & 65.8 & 33.4 \\
$^\dagger$Llama-3.1-8B-PRM800k-r & 27.4 & 69.1 & 39.3 & 14.6 & 87.2 & 25.0 & 20.1 & 64.9 & 30.7 \\ 
$^\dagger$Llama-3.1-8B-PRM800k-PDDL-r & 39.8 & 50.3 & 44.4 & 24.6 & 71.5 & 36.6 & 33.6 & 56.4 & 42.1 \\

\midrule

$^\dagger$Qwen2.5-Math-7B-MathShepherd-c & 15.2 & 86.6 & 25.9 & ~~1.4 & 99.0 & ~~2.8 & ~~4.0 & 97.8 & ~~7.7 \\ 
$^\dagger$Qwen2.5-Math-7B-MathShepherd-r & 15.1 & \textbf{87.9} & 25.8 & ~~1.4 & \textbf{99.5} & ~~2.8 & ~~4.3 & \textbf{98.1} & ~~8.3 \\
$^\dagger$Qwen2.5-Math-7B-MathShepherd-PDDL-r & 31.9 & 67.1 & 43.2 & ~~7.2 & 94.0 & 13.3 & 21.0 & 85.0 & 33.6 \\

\bottomrule

\end{tabular}
}
\caption{Complete results on MR-Ben (Math, Biology and Physics domains). $^\dagger$ indicates PRMs we trained; best results in \textbf{bold}.}
\label{tab:mrben_results_full_local_avg_2}
\end{table*}

\begin{table*}[ht]
\centering
\resizebox{0.9\textwidth}{!}{
\begin{tabular}{l | c c c | c c c | c c c }

\toprule

Model
& \multicolumn{3}{c|}{\textbf{Medicine}} 
& \multicolumn{3}{c|}{\textbf{Chemistry}}
& \multicolumn{3}{c}{\textbf{Logic}}
\\
& \textbf{Error} & \textbf{Correct} & \textbf{F1}
& \textbf{Error} & \textbf{Correct} & \textbf{F1}
& \textbf{Error} & \textbf{Correct} & \textbf{F1}
\\

\midrule

Skywork-PRM-7B & 19.1 & ~~7.0 & 10.2 & 29.0 & 25.4 & 27.1 & \textbf{26.6} & ~~8.7 & 13.1 \\

\midrule

Qwen2.5-Math-PRM-7B & 14.0 & 73.2 & 23.5 & 26.8 & 75.1 & 39.5 & 10.9 & 73.1 & 19.0 \\
$^\dagger$Qwen2.5-Math-PRM-7B-PDDL-r & 16.8 & 64.6 & 26.7 & \textbf{31.5} & 66.2 & \textbf{42.7} & 14.9 & 60.9 & 24.0 \\

\midrule

Qwen2.5-Math-7B-PRM800k & 10.5 & 88.7 & 18.8 & 21.5 & 73.6 & 33.2 & ~~4.7 & 77.9 & 8.8 \\ 
$^\dagger$Qwen2.5-Math-7B-PRM800k-r & 15.4 & 53.7 & 23.9 & 20.2 & 56.3 & 29.7 & 14.6 & 47.8 & 22.4 \\
$^\dagger$Qwen2.5-Math-7B-PRM800k-PDDL-r & 23.8 & 40.5 & \textbf{30.0} & 29.4 & 52.1 & 37.6 & 21.4 & 38.8 & \textbf{27.6} \\
$^\dagger$Qwen2.5-Math-7B-PDDL-r & ~~6.1 & 72.8 & 11.3 & ~~6.9 & 70.2 & 12.5 & 5.7 & 66.6 & 10.5 \\

\midrule
$^\dagger$Llama-3.1-8B-PRM800k-c & 15.8 & 70.8 & 25.8 & 16.1 & 74.1 & 26.4 & 11.6 & 74.1 & 20.1 \\
$^\dagger$Llama-3.1-8B-PRM800k-r & 14.0 & 84.4 & 24.0 & 16.7 & 73.6 & 27.3 & 13.4 & 69.5 & 22.4 \\
$^\dagger$Llama-3.1-8B-PRM800k-PDDL-r & 
\textbf{24.9} & 61.9 & 35.5 & 26.4 & 58.6 & 36.4 & 19.3 & 51.3 & 28.1 \\

\midrule

$^\dagger$Qwen2.5-Math-7B-MathShepherd-c & ~~0.9 & 97.7 & ~~1.7 & ~~5.2 & \textbf{96.9} & ~~9.8 & ~~0.7 & 98.2 & ~~1.4 \\
$^\dagger$Qwen2.5-Math-7B-MathShepherd-r & ~~1.1 & \textbf{98.8} & ~~2.1 & ~~4.9 & 96.3 & ~~9.4 & ~~0.3 & \textbf{99.5} & ~~0.6 \\
$^\dagger$Qwen2.5-Math-7B-MathShepherd-PDDL-r & ~~4.7 & 89.5 & ~~9.0 & 11.6 & 90.8 & 20.6 & ~~8.7 & 77.1 & 15.6 \\

\bottomrule

\end{tabular}
}
\caption{Complete results on MR-Ben (Medicine, Chemistry and Logic domains). $^\dagger$ indicates PRMs we trained; best results in \textbf{bold}.}
\label{tab:mrben_results_full_local_avg_3}
\end{table*}

\end{document}